\documentclass[runningheads]{llncs}
\usepackage[T1]{fontenc}
\usepackage{graphicx}

\usepackage{multirow}
\usepackage{amsmath,amssymb}
\usepackage{graphicx}
\usepackage{booktabs}
\usepackage{hyperref}
\usepackage{xspace}
\usepackage[breakable]{tcolorbox}

\newcommand{\LLMs}[0]{\textsc{LLMs}\xspace}

\newcommand{\VLMs}[0]{\textsc{VLMs}\xspace}
\newcommand{\VLM}[0]{\textsc{VLM}\xspace}

\newcommand{\GPT}[0]{\textsc{GPT\--5.5}\xspace}
\newcommand{\Molmo}[0]{\textsc{Molmo2}\xspace}

\newcommand{\Qwen}[0]{\textsc{Qwen\-3-VL}\xspace}

\usepackage{tikz}
\usetikzlibrary{arrows,automata,backgrounds,calc,chains,external,fit,graphs,patterns,positioning,shapes,shapes.geometric}

\tikzset{
    >=stealth',
    punkt/.style={
           rectangle,
           rounded corners,
           draw=black, very thick,
           text width=10em,
           minimum height=2em,
           text centered},
    pil/.style={
           ->,
           thick,
           shorten <=2pt,
           shorten >=2pt,},
    ionode/.style={
           rectangle,
           rounded corners,
           text width=12em,
           minimum height=4em,
           text centered},
}

\tikzset{fancy/.style={rectangle,
		rounded corners=1mm,
		ultra thin,
		draw=white,
		top color=white,
		bottom color=black!20,
		draw}}

\definecolor{highClr}{rgb}{1.0, 1.0, 0.0}

\colorlet{edgeClr}{orange!80!black}
\tikzset{sandEdge/.style={
		>=stealth,
		shorten >=1pt,
		thick,
		bend left,
		text=black,
		edgeClr,
	}}

\tikzset{fadedEdge/.style={
		->,
		>=stealth,
		shorten >=1pt,
		thick,
		edgeClr!20,
	}}

\tikzset{weigthLabel/.style={
		text=black,
		sloped,
		midway,
		}}

\tikzset{fadedWeigth/.style={
		text=lightgray!40,
		sloped,
		midway,
		anchor=south,
		}}

\tikzset{blueVertex/.style={
		rectangle,minimum size=6mm,rounded corners=3mm,
		top color=white,bottom color=blue!35!cyan!25!,
		font=\ttfamily,
		text=black,
	}}

\tikzset{blueVertexG/.style={
		rectangle,minimum size=6mm,rounded corners=3mm,
		top color=white,bottom color=blue!35!cyan!25!,
		font=\ttfamily,
		text=black,
		draw=green,
		thick,
	},
	blueVertexY/.style={
		rectangle,minimum size=6mm,rounded corners=3mm,
		top color=white,bottom color=blue!35!cyan!25!,
		font=\ttfamily,
		text=black,
		draw=yellow,
		thick,
	},
	blueVertexO/.style={
		rectangle,minimum size=6mm,rounded corners=3mm,
		top color=white,bottom color=blue!35!cyan!25!,
		font=\ttfamily,
		text=black,
		draw=orange,
		thick,
	}}

\colorlet{noteClr}{lightgray!30!white!50}

\tikzset{noteBckg/.style={
		rounded corners=8pt,fill=noteClr,
	},
	noteStl/.style={
		font=\scriptsize,
		align=center,
		text=black
	}}

\begin{document}
\title{MaxSAT-Based Feedback for Guiding Vision-Language Models in Sudoku}
%
%
\author{Pedro Orvalho\inst{1}\orcidID{0000-0002-7407-5967} \and 
Guillem Alenyà\inst{2}\orcidID{0000-0002-6018-154X} \and
Felip Manyà\inst{1}\orcidID{0000-0002-8366-1458}}

\authorrunning{P. Orvalho, et al.}


\institute{\textsuperscript{1}~Artificial Intelligence Research Institute~(IIIA), Consejo Superior de Investigaciones Científicas~(CSIC), Barcelona, Catalonia, Spain \email{pedro.orvalho@iiia.csic.es}\\
\textsuperscript{2}~Institut de Robòtica i Informàtica Industrial~(IRI-CSIC-UPC), Barcelona, Spain}
\maketitle              
\begin{abstract}
\emph{Vision--Language Models}~(\VLMs) have recently demonstrated promising performance on structured visual reasoning tasks, including grid-based puzzles. However, despite strong perceptual capabilities, these models lack explicit mechanisms for enforcing logical consistency and frequently generate assignments that violate underlying constraints.
In this paper, we propose a neuro-symbolic approach that integrates formal constraint reasoning into the \VLM solving process via a \emph{Maximum Satisfiability}~(MaxSAT) oracle. Rather than computing solutions directly, the symbolic component acts as a consistency validator and refinement engine. Candidate placements generated by the \VLM are encoded as soft clauses in a partial MaxSAT formulation, while Sudoku constraints remain hard clauses. When inconsistencies arise, the MaxSAT solver identifies a largest mutually consistent subset of assignments, which is then translated into structured textual and visual feedback to guide subsequent refinements.
We evaluate our approach on a Sudoku dataset across multiple open-source and closed-access \VLMs. Results show that MaxSAT-based feedback improves logical consistency and increases the number of solved instances, particularly in full-board refinement mode. These findings demonstrate that symbolic optimisation can enhance the reliability of vision-language~reasoning.

\end{abstract}

\section{Introduction}

Recent advances in \emph{Vision--Language Models}~(\VLMs) have enabled systems to solve structured visual reasoning tasks directly from images~\cite{xumuslr,shojaee2025illusion}, including grid-based logic puzzles such as \emph{Sudoku}~\cite{sudoku-wiki}. Despite promising empirical performance, these models lack explicit mechanisms for enforcing logical consistency and often rely on pattern recognition rather than principled constraint reasoning~\cite{shojaee2025illusion}. Consequently, their outputs may violate structural constraints or exhibit unstable behaviour when attempting to refine incorrect~solutions.

In contrast, Constraint Programming~(CP)~\cite{handbook-constraint-programming} and Satisfiability~(SAT)-based methods~\cite{biere2009handbook} provide formal guarantees of soundness and completeness with respect to the underlying symbolic model. Sudoku admits a well-established SAT encoding~\cite{DBLP:conf/isaim/LynceO06}, where each puzzle has \emph{exactly one} valid solution that corresponds to a satisfying assignment of a Boolean formula.
More generally, constraint optimisation methods, such as \emph{Maximum Satisfiability}~(MaxSAT)~\cite{maxsat-handbook-sat,li2009maxsat}, enable reasoning not only about feasibility but also about partial consistency and solution refinement. These symbolic techniques offer correctness guarantees but assume that the underlying problem has already been translated into an exact symbolic representation. In contrast, \VLMs operate directly on visual inputs, interpreting potentially ambiguous images and generating candidate symbolic assignments. This complementary division of labour naturally motivates a neuro-symbolic approach that combines neural perception with symbolic reasoning.
This raises the following research question:
\emph{Can formal constraint optimisation guide Vision--Language Models in solving visual logic puzzles such as Sudoku?}

We use Sudoku not because it is difficult for symbolic methods, but because it provides a controlled benchmark in which perception errors and logical inconsistencies can be studied independently. Although Sudoku is efficiently solvable once an exact symbolic representation is available, our objective is to investigate whether symbolic optimisation can improve the reliability of general-purpose \VLMs operating directly on visual inputs. This controlled setting isolates the benefits of symbolic feedback before extending the approach to more complex visual reasoning tasks.

Recent neuro-symbolic approaches demonstrate that combining neural models with formal solvers can improve reliability in reasoning-intensive \linebreak tasks~\cite{pan2023logic,ye2023satlm,shi2025constraintllm,aaai25-Orvalho}. However, most prior work focuses on text-based reasoning or planning domains. The integration of vision-language models with constraint optimisation techniques for structured visual problems remains~largely~unexplored.

In this paper, we introduce a hybrid neuro-symbolic approach for Sudoku solving. A \VLM generates candidate placements from the puzzle image, while a \emph{MaxSAT oracle} validates and refines these proposals using a partial MaxSAT formulation. Sudoku constraints are encoded as hard clauses, and model-generated placements are encoded as soft clauses. Rather than computing a solution outright, the MaxSAT solver selects the largest mutually consistent subset of the proposed assignments. This subset is then translated into structured textual and visual feedback, guiding subsequent refinements by the model.
The resulting architecture establishes a clear division of labour: the neural component performs perception and heuristic proposal generation, whereas the symbolic component enforces logical consistency through optimisation. Consequently, every accepted placement is formally verified against the underlying constraint model. Our goal is not to outperform symbolic solvers, but to study whether symbolic optimisation can improve the reliability of general-purpose \VLMs.

Our empirical evaluation on a benchmark Sudoku dataset~\cite{sudoku-dataset-huggingface} shows that integrating partial MaxSAT refinement into the \VLM solving loop substantially improves logical consistency, solution completeness, and solve rates across both open-source and closed-access \VLMs. The strongest gains are observed in full-board solving scenarios, where symbolic refinement effectively repairs globally coherent but partially inconsistent neural predictions, enabling significantly more puzzles to be solved correctly, including more challenging instances.

In summary, this paper makes the following contributions:

\begin{itemize}
  \item A MaxSAT-based neuro-symbolic framework for validating and refining \emph{Vision--Language Models}~(\VLM)-generated Sudoku solutions.
  \item The integration of partial MaxSAT into the \VLM solving loop, where model-generated placements are treated as soft constraints and refined through constraint~optimisation.
  \item An empirical evaluation across multiple open-source and closed-access \VLMs demonstrating that MaxSAT-based refinement improves logical consistency and increases significantly the number of solved instances.
\end{itemize}

\section{Preliminaries}
\label{sec:prelim}

The \emph{Boolean Satisfiability}~(SAT) problem is the canonical decision problem of propositional logic~\cite{biere2009handbook}. A \emph{literal} is either a propositional variable $x_i$ or its negation $\neg x_i$. A formula is in \emph{Conjunctive Normal Form}~(CNF) if it is expressed as a conjunction of clauses, where each clause is a disjunction of literals.
Given a CNF formula $\phi$, the SAT problem consists of determining whether there exists a truth assignment to its variables that satisfies all clauses of $\phi$, or whether $\phi$ is unsatisfiable.
The \emph{Maximum Satisfiability}~(MaxSAT) problem extends SAT with an optimisation objective~\cite{maxsat-handbook-sat,li2009maxsat}. Given a CNF formula $\phi$, the goal is to compute an assignment that maximises the number of satisfied clauses.
In the \emph{partial MaxSAT} setting, the formula $\phi$ is partitioned into a set of hard clauses $\phi_h$ and a set of soft clauses $\phi_s$. The objective is to find an assignment that satisfies all clauses in $\phi_h$ while minimising the number of unsatisfied clauses in $\phi_s$. In the \emph{partial MaxSAT} variant, the objective becomes minimising the total number of unsatisfied soft clauses.

\begin{example}[Partial MaxSAT]
  Consider the partial MaxSAT formula $\phi = (\phi_h, \phi_s)$, where
  $\phi_h = \{ (x_1 \vee x_2), (\neg x_2 \vee x_3) \}$ and
  $\phi_s = \{ (\neg x_1), (\neg x_3) \}$.
  The assignment $\{ (x_1, 1), (x_2, 0), (x_3, 0) \}$ satisfies all hard clauses in $\phi_h$ and violates only the soft clause $(\neg x_1)$, incurring a total cost of~1. Hence, it is an optimal solution.
\end{example}

\section{Sudoku as a MaxSAT Problem}
\label{sec:sat-encoding}

Let $n = 9$ and let $r,c,d \in \{1,\dots,n\}$. We introduce Boolean variables $X_{r,c,d} \in \{0,1\}$, where $X_{r,c,d} = 1$ denotes that digit $d$ is assigned to cell $(r,c)$. 

Thus, we can easily encode the Sudoku problem as a constraint satisfaction problem, following previously used encoding~\cite{DBLP:conf/isaim/LynceO06}, by encoding it as a Satisfiability~(SAT) problem. Using the following constraints:
\begin{itemize}
    \item \textbf{Cell Constraints.} Each cell must contain exactly one digit: $\forall r,c \in \{1,\dots,n\}: \sum_{d=1}^{n} X_{r,c,d} = 1$.
    \item \textbf{Row Constraints.} Each digit must appear exactly once in every row: $\forall r,d \in \{1,\dots,n\}: \sum_{c=1}^{n} X_{r,c,d} = 1$.
    \item \textbf{Column Constraints.} Each digit must appear exactly once in every column: $\forall c,d \in \{1,\dots,n\}: \sum_{r=1}^{n} X_{r,c,d} = 1$.
    \item \textbf{Subgrid Constraints.} Each digit must appear exactly once in every $sqrt(n) \times sqrt(n)$ subgrid: $\sum_{r \in R_b} \sum_{c \in C_b} X_{r,c,d} = 1 \qquad \forall d \in \{1,\dots,n\}, \; \forall b \in \{1,\dots,9\}$, where $(R_b, C_b)$ denotes the row and column indices of subgrid $b$.
\end{itemize}

\textbf{CNF Encoding.}
Each exactly-one constraint is encoded in \emph{Conjunctive Normal Form~(CNF)} using a standard pairwise encoding: 
(i) an at-least-one clause enforcing that at least one literal in the group is true, and 
(ii) pairwise at-most-one clauses preventing two literals from being simultaneously true.
The resulting CNF formula constitutes the hard constraint set of the Sudoku instance and can be solved using a SAT solver.

\textbf{MaxSAT Formulation.}
In our setting, Vision--Language Models~(\VLMs)-proposed placements (i.e., plays) are encoded as soft unit clauses in a partial Maximum Satisfiability~(MaxSAT) formulation. The objective is to satisfy all Sudoku constraints (hard clauses) while maximising the number of consistent model-proposed placements (soft clauses). This formulation enables refinement of candidate solutions via MaxSAT optimisation.

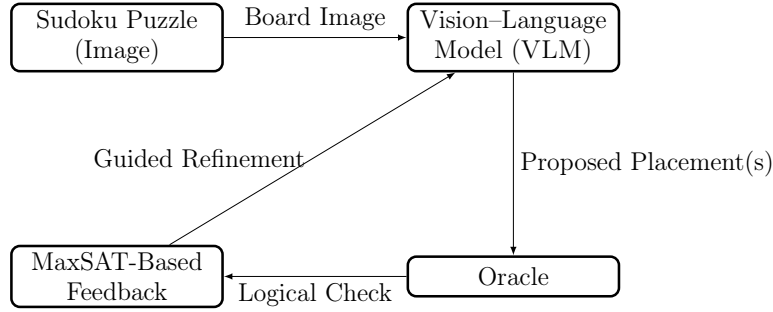
\begin{figure*}[t!]
\centering
\resizebox{0.85\columnwidth}{!}{
\begin{tikzpicture}[
    node distance=3cm and 3cm,
    every node/.style={font=\fontsize{12}{14}\selectfont},
    >=latex
]

\node[punkt] (puzzle) {Sudoku Puzzle\\(Image)};
\node[punkt, right=of puzzle] (vlm) {Vision--Language Model (\VLM)};

\node[punkt, below=of vlm] (oracle) {Oracle};
\node[punkt, left=of oracle] (feedback) {MaxSAT-Based Feedback};

\draw[->] (puzzle) -- node[above]{Board Image} (vlm);

    \draw[->] (vlm) -- node[right]{Proposed Placement(s)} (oracle);

\draw[->] (oracle) -- node[below]{Logical Check} (feedback);

\draw[->] (feedback) -- node[left]{Guided Refinement} (vlm);

\end{tikzpicture}
}
\caption{Neuro-symbolic interaction loop for Sudoku solving. The \VLM proposes candidate placements, which are validated by a MaxSAT oracle. When conflicts are detected, constraint-derived feedback is highlighted in the puzzle and returned to guide~subsequent~refinements.}
\label{fig:approach}
\end{figure*}

\section{Neuro-Symbolic Interaction Loop}
\label{sec:interaction}

Figure~\ref{fig:approach} illustrates the neuro-symbolic interaction loop underlying our approach. The framework combines a Vision--Language Model~(\VLM) with a symbolic partial Maximum Satisfiability~(MaxSAT) oracle to iteratively construct logically consistent Sudoku assignments.

Let $\mathcal{P}$ denote a Sudoku puzzle represented as an image, and let $\mathcal{X} = \{X_{r,c,d} \mid r,c,d \in \{1,\dots,9\}\}$ be the set of Boolean variables introduced in Section~\ref{sec:sat-encoding}, where $X_{r,c,d}=1$ indicates that digit $d$ is assigned to cell $(r,c)$.

Given an image of the Sudoku puzzle, the \VLM generates a set of candidate assignments $\hat{A} = \{(r,c,d)\}$, where each assignment corresponds to placing digit $d$ in cell $(r,c)$ and is associated with the Boolean literal $X_{r,c,d}$ in the symbolic encoding.
The symbolic component then constructs a partial MaxSAT instance $\Phi = (\Phi_h,\Phi_s)$, where $\Phi_h$ contains the hard clauses encoding all Sudoku constraints, and $\Phi_s$ contains soft unit clauses corresponding to the assignments proposed by the \VLM.
More precisely, for each proposed placement $(r,c,d)\in \hat{A}$, the oracle introduces the soft clause $(X_{r,c,d})$.

The MaxSAT solver then computes an assignment satisfying all hard clauses while maximising the number of satisfied soft clauses. If all soft clauses are satisfied, the proposal is logically consistent with the Sudoku constraints. Otherwise, the set of unsatisfied soft clauses corresponds to assignments rejected by the MaxSAT solver, which are interpreted as logically inconsistent placements.

These inconsistencies are translated into structured feedback consisting of: (i) textual descriptions identifying inconsistent placements; and (ii) visual annotations highlighting conflicting cells in the puzzle image.
Conditioned on this feedback, the \VLM generates a refined assignment proposal, and the interaction loop repeats until either: (i) a complete and logically consistent solution is obtained; or (ii) a predefined termination condition is reached.

Importantly, logical consistency checking and optimisation are delegated entirely to the symbolic oracle. The \VLM is therefore responsible only for perceptual interpretation and candidate generation, while formal constraint satisfaction is guaranteed by the MaxSAT component.

We evaluate two interaction protocols within this framework: (i) iterative single-placement solving; and (ii) full-board refinement. 
All prompts used in our experiments are provided in Appendix~\ref{sec:prompts}.

\subsection{Iterative Sudoku Solving}

In iterative mode, the \VLM proposes a single placement at each interaction step. Let $a_t = (r_t,c_t,d_t)$ denote the placement proposed at iteration $t$.
The proposed assignment is encoded as a soft unit clause and added to the partial MaxSAT instance. The solver then computes an optimal assignment satisfying all hard Sudoku constraints while maximising agreement with the proposed placements.
If $a_t$ belongs to the optimal MaxSAT solution, the placement is accepted and permanently added to the board state. Otherwise, the placement is rejected and returned to the \VLM as corrective feedback. The process repeats until either a valid Sudoku solution is obtained or a termination condition is reached.
This interaction protocol can therefore be interpreted as an optimisation-guided sequential decision process in which symbolic consistency checking constrains the generation behaviour of the \VLM.

If we assume that the \VLM always returns exactly one placement in this setting, then a standard SAT-based verifier would be sufficient to check whether that placement is consistent with the Sudoku constraints. However, we use a MaxSAT solver to retain the same formulation when the \VLM outputs more than one placement, as this allows us to identify an optimal correction in the sense of minimising the number of logically incorrect proposed placements. In other words, MaxSAT generalises the SAT-based check: when a single placement is proposed, both approaches coincide, although the SAT-based approach should be more efficient; whereas when multiple placements are produced, MaxSAT returns a minimum set of conflicting assignments to be rejected, thereby providing a more informative and robust feedback signal~to~the~\VLM.

\subsection{Full-Board Refinement}

In full-board mode, the \VLM proposes a complete Sudoku assignment in a single interaction. All predicted placements are encoded as soft unit clauses in the partial MaxSAT formulation.
Given the proposed assignment set $\hat{A}$, the MaxSAT solver computes a largest mutually consistent subset $\hat{A}^* \subseteq \hat{A}$, such that all Sudoku constraints remain satisfied.
Placements in $\hat{A} \setminus \hat{A}^*$ are identified as inconsistent assignments. These conflicts are translated into structured textual and visual feedback and returned to the \VLM, which then generates a refined full-board proposal.
The refinement procedure is repeated for a bounded number of iterations or until a complete valid Sudoku solution is obtained.

\section{Experiments}
\label{sec:results}

The goal of our experimental evaluation is to assess the reasoning capabilities of Vision--Language Models~(\VLMs) on Sudoku puzzles, and to determine whether Maximum Satisfiability (MaxSAT)-based feedback can improve their ability to construct logically consistent solutions. In particular, we study whether constraint-guided feedback enables \VLMs to correct invalid placements and progressively converge toward a complete solution.

Our evaluation is structured around the following research questions (RQs):
\\
\textbf{RQ1.} To what extent do \VLMs produce placements that satisfy Sudoku constraints without symbolic assistance?
\\
\textbf{RQ2.} Does MaxSAT-based feedback improve the ability of \VLMs to correct invalid moves and reach a complete and logically correct solution?
\\
\textbf{RQ3.} How do open-source and closed-access \VLMs differ in their ability to solve Sudoku puzzles under iterative constraint-guided interaction?
\\
\textbf{RQ4.} How does \VLM performance vary across different levels of puzzle~difficulty?

\subsection{Experimental Setup}

All experiments were conducted on a machine equipped with an NVIDIA A100-SXM4 GPU 40GB and 128GB of RAM. Each interaction (i.e., model play) was limited to 5 minutes, and the total time limit per puzzle was capped at 30 minutes. These limits ensure a fair and uniform comparison across models while preventing excessively long interaction loops.
For symbolic reasoning and refinement, we used the RC2 MaxSAT solver~\cite{imms19-RC2} from the \texttt{PySAT} toolkit~\cite{imms18-PySAT}. 

The solving process for a given Sudoku puzzle is terminated under one of the following conditions: (1) the puzzle is successfully completed; (2) the model proposes the same incorrect placement ten times, indicating repetitive incorrect behavior without progress; or (3) the 30-minute puzzle time limit is reached. 
All models are evaluated under identical interaction protocols and time constraints to ensure comparability. 
All prompts used in our experiments are provided in Appendix~\ref{sec:prompts}.

\subsection{Evaluation Dataset}

We evaluate our approach on the Sudoku dataset available on HuggingFace~\cite{sudoku-dataset-huggingface}, which contains $9\times9$ puzzles annotated with difficulty levels and corresponding ground-truth solutions. 
We restrict our evaluation to instances labelled with difficulty levels $0$ and $1$, representing easier and harder puzzles. Using a fixed random seed ($42$), we uniformly sample $100$ puzzles from each level without replacement, resulting in a total of $200$ instances. 
All models are evaluated on the same fixed subset to ensure reproducibility and a fair comparison~across~models. 
Importantly, the MaxSAT solver RC2 solves all the Sudoku instances in our dataset in a few seconds. This highlights that the challenge is not solving the constraint satisfaction problem itself, but guiding the \VLM towards~consistent~solutions.

\subsection{Vision-Language Models~(\VLMs)}

In our evaluation, we consider three \VLMs: two open-source models and one closed-access model.
For the open-source models, we select \VLMs available on Hugging Face~\cite{huggingface}: Alibaba’s \Qwen~\cite{yang2025qwen3}~(32B, 2025), and AllenAI’s \linebreak \Molmo~\cite{clark2026molmo2}~(4B, 2026).
For the closed-access model, we evaluate OpenAI’s \GPT~\cite{chatgpt5.5}~(2026), as this is among the most widely adopted state-of-the-art proprietary models.
To ensure consistency across experiments, all model interactions are conducted without any prior context, only the information regarding the solving process of a given puzzle is retained between queries, and the temperature of all models was~set~to~zero.

\begin{table}[t]
\centering
\caption{Number of solved instances and average completeness (\%) overall and by difficulty. Best results per approach are highlighted in bold.}
\label{tab:completeness_one_table_by_difficulty_compact}

\setlength{\tabcolsep}{3pt} 
\renewcommand{\arraystretch}{1.1}

\resizebox{\columnwidth}{!}{%
\begin{tabular}{llrrrrrrr}
\toprule
\multirow{2}{*}{\textbf{Approach}} & \multirow{2}{*}{\textbf{VLM}}
& \multicolumn{3}{c}{\textbf{All}} & \multicolumn{2}{c}{\textbf{Diff.\ 0}} & \multicolumn{2}{c}{\textbf{Diff.\ 1}} \\
\cmidrule(lr){3-5}\cmidrule(lr){6-7}\cmidrule(lr){8-9}
& & \textbf{\# Solved} &  \textbf{\% Solved} &  \textbf{AC\ (\%)} & \textbf{\# Solved} & \textbf{AC\ (\%)} & \textbf{\# Solved} & \textbf{AC\ (\%)} \\
\midrule
\midrule
\multirow{3}{*}{\texttt{\textbf{Step-by-Step (Validity Only)}}}
  & \GPT     & \textbf{21} & \textbf{10.5\%} & \textbf{44.8} & 16 & 63.9 & 5 & 25.6 \\
  & \Molmo    & 5 & 2.5\% & 35.9 & 4 & 49.4 & 1 & 22.3 \\
  & \Qwen     & 2 & 1.0\% & 34.7 & 2 & 47.3 & 0 & 22.1 \\
\midrule
\multirow{3}{*}{\texttt{\textbf{Step-by-Step (MaxSAT Feedback)}}}
  & \GPT      & \textbf{44} & \textbf{22.0\%} & \textbf{52.2} & 33 & 72.8 & 11 & 31.6 \\
  & \Molmo   & 11 & 5.5\% & 41.6 & 7 & 55.1 & 4 & 28.0 \\
  & \Qwen    & 6 & 3.0\% & 36.5 & 5 & 50.2 & 1 & 22.8 \\
\midrule
\midrule
\multirow{3}{*}{\texttt{\textbf{Full Board (Validity Only)}}}
  & \GPT      & \textbf{45} & \textbf{22.5\%} & \textbf{72.0} & 35 & 81.7 & 10 & 62.2 \\
  & \Molmo    & 17 & 8.5\% & 45.4 & 12 & 56.3 & 5 & 34.5 \\
  & \Qwen     & 10 & 5.0\% & 38.9 & 7 & 51.9 & 3 & 25.8 \\
\midrule
\multirow{3}{*}{\texttt{\textbf{Full Board (MaxSAT Feedback)}}}
  & \GPT      & \textbf{73} & \textbf{36.5\%} & \textbf{80.7} & 52 & 84.6 & 21 & 76.7 \\
  & \Molmo    & 28 & 14.0\% & 51.6 & 19 & 64.5 & 9 & 38.6 \\
  & \Qwen     & 13 & 6.5\%  & 42.6 & 8 & 56.3 & 5 & 28.9 \\
\bottomrule
\end{tabular}%
}
\label{tab:results}
\end{table}

\subsection{Results}
\label{sec:results}

Table~\ref{tab:results} reports the number of solved instances and the average completeness~(AC) of the final board, both overall and divided by difficulty. Average completeness~(AC) is defined as the proportion of cells in the final board that match the ground-truth solution, expressed as a percentage. As described in Section~\ref{sec:interaction}, we compare two interaction approaches: \texttt{Step-by-Step} (one placement per interaction) and \texttt{Full Board} (a complete solution proposed in a single interaction). Each approach is evaluated under two feedback regimes: validity-only textual feedback and MaxSAT-based feedback providing structured textual and visual guidance.
Across all configurations, \GPT consistently achieves the strongest performance in both solve rate and completeness. The open-source models, \Molmo and \Qwen, achieve substantially lower scores overall, indicating greater difficulty in producing globally consistent Sudoku assignments from visual input alone.

\textbf{Step-by-step solving.}
In the \texttt{Step-by-Step} setting, MaxSAT-based feedback substantially improves performance across all models, particularly for \GPT. Under validity-only feedback, \GPT solves 21 puzzles overall with an average completeness of $44.8\%$. With MaxSAT feedback, solved instances more than double to 44, while AC increases to $52.2\%$. Improvements are observed for both difficulty levels, with difficulty~0 solved instances increasing from 16 to 33 and difficulty~1 solved instances increasing from 5 to 11.

The open-source models also benefit from refinement, although to a lesser extent. \Molmo improves from 5 solved instances to 11 and increases AC from $35.9\%$ to $41.6\%$. \Qwen improves from 2 to 6 solved puzzles, with a smaller increase in completeness. These results indicate that symbolic feedback can partially compensate for inconsistent local reasoning during the iterative placement generation process.

\textbf{Full-board solving.}
The strongest overall results are achieved in the \texttt{Full Board (MaxSAT Feedback)} configuration. In this setting, \GPT solves 73 puzzles overall, corresponding to a $36.5\%$ solve rate, while achieving an average completeness of $80.7\%$. This includes 52 solved difficulty~0 puzzles and 21 solved difficulty~1 puzzles, demonstrating that MaxSAT-based refinement remains effective even on harder instances when applied to globally structured predictions.

Compared to \texttt{Full Board (Validity Only)}, where \GPT solves 45 puzzles with $72.0\%$ AC, MaxSAT refinement produces substantial gains in both solve rate and completeness. Notably, the improvement on difficulty~1 puzzles is particularly strong, with AC increasing from $62.2\%$ to $76.7\%$ and solved instances more than doubling from 10 to 21.

The open-source models also show improvements under full-board refinement. \Molmo increases from 17 to 28 solved instances overall and improves AC from $45.4\%$ to $51.6\%$. \Qwen improves more modestly, increasing from 10 to 13 solved instances and achieving $42.6\%$ AC under refinement.

\textbf{Effect of puzzle difficulty.}
Performance consistently declines from difficulty~0 to difficulty~1 across all models and interaction settings, reflecting the increased complexity of harder Sudoku instances. Nevertheless, MaxSAT-based refinement significantly reduces this performance gap, especially for \GPT in the full-board setting. While harder puzzles remain more challenging overall, symbolic refinement enables the recovery of many near-consistent assignments that would otherwise remain invalid.

\subsection{Discussion}
\label{sec:discussion}

The results in Table~\ref{tab:results} allow us to address the RQs introduced earlier.

\textbf{Answering RQ1.}
Validity-only feedback reveals that \VLMs frequently generate assignments that violate Sudoku constraints or remain incomplete. Although \GPT achieves moderate success without symbolic assistance, particularly in the full-board setting, performance remains limited overall, especially on more difficult puzzles. The open-source models perform substantially worse under validity-only feedback, with low solve rates and lower completeness scores. These findings confirm that, without explicit constraint-based refinement, \VLM outputs often lack sufficient logical consistency for reliable Sudoku solving.

\textbf{Answering RQ2.}
MaxSAT-based refinement consistently improves both solve rate and completeness across all interaction settings and models. The largest improvements are observed for \GPT in the full-board setting, where solved instances increase from 45 to 73 and average completeness rises from $72.0\%$ to $80.7\%$. Significant gains are also observed on difficulty~1 puzzles, where solved instances more than double.

In the step-by-step setting, MaxSAT feedback also substantially improves performance, increasing solved instances for \GPT from 21 to 44. Although improvements for \Molmo and \Qwen are smaller, both models still benefit from symbolic refinement. These findings demonstrate that MaxSAT-based guidance effectively repairs partially inconsistent neural predictions by identifying a largest mutually consistent subset of placements. The results further suggest that symbolic refinement is especially effective when the \VLM generates globally coherent but imperfect candidate solutions.

\textbf{Answering RQ3.}
A clear performance gap remains between the proprietary \GPT model and the open-source \VLMs. Across all configurations, \GPT achieves substantially higher solve rates and completeness scores, and it also benefits more strongly from MaxSAT refinement. In contrast, \Molmo achieves moderate gains under refinement, while \Qwen shows comparatively limited improvement.
These findings suggest that the effectiveness of symbolic refinement depends strongly on the quality of the initial neural proposal. When the initial assignments are closer to global consistency, MaxSAT refinement is more effective at recovering a fully valid solution.
At the same time, these performance differences should also be considered in terms of \emph{accessibility} and \emph{cost}. While \Molmo and \Qwen are openly available models, \GPT requires paid API access. Consequently, the substantially higher performance achieved by the proprietary model comes with a non-trivial financial and computational cost trade-off.

\textbf{Answering RQ4.}
Puzzle difficulty has a substantial impact on performance across all models and configurations. Difficulty~1 puzzles consistently yield lower solve rates and completeness scores than difficulty~0 puzzles, highlighting the increased reasoning demands imposed by harder instances.
However, MaxSAT refinement significantly mitigates this degradation, particularly for \GPT in the full-board setting. The strong improvement on harder puzzles indicates that symbolic optimisation can compensate for some limitations in neural reasoning by enforcing global consistency constraints during iterative refinement.

Overall, these findings strengthen the central claim of this work: integrating partial MaxSAT into the \VLM solving loop significantly improves logical consistency, robustness, and solution quality. While symbolic refinement is not a complete remedy, it proves particularly effective when neural models generate near-consistent global assignments that require structured correction.

Note that a standard Sudoku puzzle consists of a $9 \times 9$ grid with a \emph{unique valid solution} satisfying all row, column and subgrid constraints. Finding a solution to an $n \times n$ Sudoku puzzle is NP-complete~\cite{yato2003complexity}. The complexity of this task helps explain the observed difficulty of \VLMs in consistently producing complete and valid assignments, and further motivates the need for constraint-based refinement to support robust and explainable reasoning.

\section{Related Work}
\label{sec:related}

\emph{Sudoku as a Constraint Satisfaction Problem (CSP).}
Sudoku is a canonical benchmark for constraint satisfaction and has been extensively studied within the CP and SAT communities~\cite{DBLP:conf/isaim/LynceO06,DBLP:journals/corr/abs-2511-10436,DBLP:journals/constraints/MulambaMMG24,DBLP:conf/aaai/GunsGMBBP23}. The classical encoding~\cite{DBLP:conf/isaim/LynceO06} shows that each valid solution corresponds to a satisfying assignment of a CNF formula. In traditional SAT-based approaches, the solver computes complete solutions directly. In contrast, our work employs partial MaxSAT as a refinement mechanism within a neural reasoning loop rather than as a standalone solver.

\emph{Explainable AI for Logic Puzzles.}
In recent years, research in eXplainable Artificial Intelligence~(XAI)~\cite{DBLP:conf/ijcai/Ignatiev20,DBLP:conf/cade/IgnatievPNM18,DBLP:journals/frai/MarquesSilvaI23} has increasingly focused on logic puzzles, such as Sudoku, which serve as controlled environments for studying explainability and reasoning transparency. Recent work~\cite{DBLP:conf/ecai/0001GCG20,DBLP:journals/constraints/MulambaMMG24,DBLP:conf/aaai/GunsGMBBP23} investigates the extraction of human-readable explanations from constraint solvers and the analysis of reasoning traces in symbolic systems. These approaches aim to make symbolic reasoning processes interpretable and transparent. 
Our setting differs in that symbolic reasoning is not explained post hoc, rather, it is used proactively as a MaxSAT-based consistency filter to refine vision models' outputs.

\emph{Neuro-Symbolic Reasoning with Language Models.}
There is growing interest in combining Large Language Models (\LLMs) with formal reasoning tools to improve reliability in reasoning-intensive tasks. Such approaches have been applied to planning~\cite{kambhampati2024position,DBLP:conf/naacl/HaoCZF25,haoplanning}, arithmetic~\cite{gao2023pal}, and logical reasoning~\cite{linzebralogic,ye2023satlm,pan2023logic,shi2025constraintllm}. Notably, \texttt{SATLM}~\cite{ye2023satlm} and \texttt{LOGIC-LM}~\cite{pan2023logic} translate natural language problems into formal logical specifications that are solved by SMT or SAT engines. 
While most prior approaches focus on text-based reasoning, our work targets structured visual reasoning with Vision-Language Models (\VLMs). Furthermore, rather than translating the entire problem into a symbolic representation, we retain the neural model as the primary proposal generator and use partial Maximum Satisfiability~(MaxSAT) to refine and validate its assignments.

\emph{MaxSAT-Guided Neural Systems.}
MaxSAT has recently been explored as a mechanism for guiding or repairing neural model outputs, including applications in code generation and verification~\cite{aaai25-Orvalho,ok25-arXiv-pyVeritas}. These works demonstrate that optimisation-based filtering can improve correctness while preserving neural flexibility. Our work extends this paradigm to structured visual constraint problems, showing that MaxSAT-based refinement can enhance logical consistency in \VLM-based Sudoku solving.

Thus, prior work has established Sudoku as a classical CSP and explored neuro-symbolic reasoning mainly in text-based domains. Furthermore, prior work mainly focuses on building specialised Sudoku solvers. In contrast, our setting studies whether a general-purpose \VLM can be guided via constraint-derived feedback.
Unlike prior neuro-symbolic approaches that invoke symbolic solvers to compute complete solutions, our framework uses MaxSAT exclusively as a feedback oracle, preserving the Vision-Language Model~(\VLM) as the primary reasoning agent.
The focus is thus on the interaction mechanism rather than competing with specialised systems.

\section{Conclusion}
\label{sec:conclusion}

We propose a MaxSAT-based neuro-symbolic framework for guiding Vision--Language Models~(\VLMs) in solving structured visual reasoning tasks such as Sudoku. Rather than replacing neural models with symbolic solvers, our approach integrates partial Maximum Satisfiability~(MaxSAT) as a consistency and refinement oracle. By encoding Sudoku constraints as hard clauses and model-generated placements as soft clauses, the MaxSAT solver identifies a largest mutually consistent subset of assignments, providing structured textual and visual feedback to the \VLM. Thus, in our work, a MaxSAT solver improves reliability of \VLM outputs under structured visual reasoning.

Our empirical evaluation demonstrates that this MaxSAT-based refinement improves logical consistency and increases the number of solved instances, particularly in full-board solving scenarios. The results highlight the complementary strengths of neural perception and symbolic optimisation: while \VLMs generate candidate solutions, MaxSAT ensures formal consistency of accepted placements.

These findings reinforce the potential of neuro-symbolic systems in structured reasoning tasks. 
Although Sudoku serves as our experimental benchmark, its role is to provide a controlled setting for evaluating symbolic feedback mechanisms that can be transferred to more general visual reasoning tasks.
Future work includes extending our approach to other visual constraint satisfaction problems and exploring tighter integration between neural proposal mechanisms and symbolic optimisation layers. Overall, this work illustrates how classical constraint-based techniques can enhance the reliability of Vision-Language Models, advancing the synergy between statistical and symbolic artificial intelligence.

\begin{credits}
\subsubsection{\ackname} 
This work was supported by grants PID2022-139835NB-C21 and PID2025-171340NB-I00 funded by MCIN/\-AEI/\-10.13039/\-501100011033.
This project has received funding from the European Union’s HORIZON-MSCA-2025-PF research and innovation programme under the Marie Skłodowska-Curie  (Sherlock4Py, \break GA~No~101269051).
This project was additionally supported by the ALLIES Cofund, and has received funding from the European Union’s Horizon-MSCA-2022-COFUND-01 research and innovation programme under the Marie Skłodowska-Curie~(GA \break No~101126626). G. Alenyà acknowledges the support of EU project FlexCycle HORIZON-CL4-2024-DIGITAL-EMERGING-01-101189600.
\end{credits}

%
%

\bibliographystyle{splncs04}
\bibliography{bibfile}

\newpage

\appendix

\section{Prompt Templates}
\label{sec:prompts}

This appendix presents the prompt templates used to evaluate Vision–Language Models across the Sudoku solving configurations described in Section~\ref{sec:interaction}. The prompts were designed to be minimal, unambiguous and consistent across models.
In all experiments, each prompt consists of a puzzle image paired with a single instruction. Models are required to respond using a fixed answer format, enclosed within a dedicated \texttt{<ANSWER>} tag, to enable reliable parsing and systematic evaluation.

\subsection{Step-by-Step Prompts}

In the Step-by-Step setting, the prompting procedure consists of an initial instruction followed by two types of feedback prompts: validity-only feedback and MaxSAT-based feedback.

\begin{tcolorbox}[breakable,title=Initial Prompt]
\begin{verbatim}
You are a professional Sudoku solver.
Sudoku is a logic-based number placement puzzle played on a 
9×9 grid. The objective is to fill the grid with digits from 
1 to 9 so that eachrow, each column, and each of the nine 3×3 
subgrids contains every digit exactly once. 
We give you an image of a Sudoku board.
Provide exactly one placement you are confident about.
Output ONLY an <ANSWER>...</ANSWER> block containing 
your placement.
Answer format (exact):
<ANSWER>r{row}c{col}: {digit}</ANSWER>
ONLY output that block and NOTHING else.
Example:
<ANSWER>r1c4: 9</ANSWER>
Your answer:
\end{verbatim}
\end{tcolorbox}

\begin{tcolorbox}[breakable,title=Validity-Only Feedback Prompt]
\begin{verbatim}
FEEDBACK ON YOUR LAST PLAY:
Your placement is logically incorrect.
Try Again.
Your answer:
\end{verbatim}
\end{tcolorbox}

\begin{tcolorbox}[breakable,title=MaxSAT-Based Feedback Prompt]
\begin{verbatim}
FEEDBACK ON YOUR LAST PLAY:
Your placement is logically incorrect.
Conflicts: r1c1=..., ...
The conflicting cells are highlighted in red.
Try Again.
Your answer:
\end{verbatim}
\end{tcolorbox}

\subsection{Full Board (MaxSAT Feedback).}

In the Full Board setting, the prompting procedure consists of an initial instruction followed by two types of feedback prompts: validity-only feedback and MaxSAT-based feedback.

\begin{tcolorbox}[breakable,title=Initial Prompt]
\begin{verbatim}
You are a professional Sudoku solver.
Sudoku is a logic-based number placement puzzle played on a 
9×9 grid. The objective is to fill the grid with digits from 
1 to 9 so that eachrow, each column, and each of the nine 3×3 
subgrids contains every digit exactly once. 
We give you an image of a Sudoku board.
Provide a COMPLETE Sudoku solution.
Output ONLY an <ANSWER>...</ANSWER> block containing the 
full board.
Answer format (exact):
<ANSWER>
r{row}c{col}: {digit}
...
r{row2}c{col2}: {digit}
</ANSWER>
ONLY output that block and NOTHING else.
Example:
<ANSWER>
r1c4: 9
r2c3: 5
</ANSWER>
Your answer:
\end{verbatim}
\end{tcolorbox}

\begin{tcolorbox}[breakable,title=Validity-Only Feedback Prompt]
\begin{verbatim}
FEEDBACK ON YOUR LAST FULL SOLUTION:
Your full-board solution contains logical conflicts.
Try Again.
Your answer:
\end{verbatim}
\end{tcolorbox}

\begin{tcolorbox}[breakable,title=MaxSAT-Based Feedback Prompt]
\begin{verbatim}
FEEDBACK ON YOUR LAST FULL SOLUTION:
Your full-board solution contains logical conflicts.
Conflicting placements: r1c1=..., ...
The conflicting cells are highlighted in red.
Try Again.
Your answer:
\end{verbatim}
\end{tcolorbox}

\end{document}